\documentclass{article}

\usepackage[english]{babel}

\usepackage[letterpaper,top=2cm,bottom=2cm,left=3cm,right=3cm,marginparwidth=1.75cm]{geometry}

\usepackage{amsmath}
\usepackage{graphicx}
\usepackage[colorlinks=true, allcolors=blue]{hyperref}
\usepackage{authblk}
\usepackage{graphicx}%
\usepackage{multirow}%
\usepackage{amssymb,amsfonts}%
\usepackage{mathrsfs}%
\usepackage[title]{appendix}%
\usepackage{xcolor}%
\usepackage{textcomp}%
\usepackage{manyfoot}%
\usepackage{booktabs}%
\usepackage{algorithm}%
\usepackage{algorithmicx}%
\usepackage{algpseudocode}%
\usepackage{listings}%
\usepackage{subcaption}
\usepackage{tabularx}
\usepackage{listings}
\usepackage{array}
\usepackage{fontawesome}

\definecolor{background}{HTML}{EEEEEE}
\definecolor{delim}{RGB}{20,105,176}
\colorlet{numb}{magenta}

\lstdefinelanguage{json}{
    basicstyle=\normalfont\ttfamily,
    numbers=left,
    numberstyle=\scriptsize,
    stepnumber=1,
    numbersep=8pt,
    showstringspaces=false,
    breaklines=true,
    frame=lines,
    backgroundcolor=\color{background},
    literate=
     *{0}{{{\color{numb}0}}}{1}
      {1}{{{\color{numb}1}}}{1}
      {2}{{{\color{numb}2}}}{1}
      {3}{{{\color{numb}3}}}{1}
      {4}{{{\color{numb}4}}}{1}
      {5}{{{\color{numb}5}}}{1}
      {6}{{{\color{numb}6}}}{1}
      {7}{{{\color{numb}7}}}{1}
      {8}{{{\color{numb}8}}}{1}
      {9}{{{\color{numb}9}}}{1}
      {:}{{{\color{delim}{:}}}}{1}
      {,}{{{\color{delim}{,}}}}{1}
      {\{}{{{\color{delim}{\{}}}}{1}
      {\}}{{{\color{delim}{\}}}}}{1}
      {[}{{{\color{delim}{[}}}}{1}
      {]}{{{\color{delim}{]}}}}{1},
}

\title{Distilling Large Language Models for Matching Patients to Clinical Trials}
\author[1]{Mauro Nievas\thanks{Mauro.nievasoffidani@triomics.com}}
\author[2]{Aditya Basu\thanks{aditya.basu@triomics.in}}
\author[3]{Yanshan Wang\thanks{yanshan.wang@pitt.edu}}
\author[1]{Hrituraj Singh\thanks{hrituraj@triomics.com}}
\affil[1]{Triomics Research, San Francisco, USA}
\affil[2]{Triomics Research, Bangalore, India}
\affil[3]{University of Pittsburgh, Pittsburgh, USA}

\begin{document}
\maketitle

\begin{abstract}
The recent success of large language models (LLMs) has paved the way for their adoption in the high-stakes domain of healthcare. Specifically, the application of LLMs in patient-trial matching, which involves assessing patient eligibility against clinical trial's nuanced inclusion and exclusion criteria, has shown promise. Recent research has shown that GPT-3.5, a widely recognized LLM developed by OpenAI, can outperform existing methods with minimal 'variable engineering' by simply comparing clinical trial information against patient summaries. However, there are significant challenges associated with using closed-source proprietary LLMs like GPT-3.5 in practical healthcare applications, such as cost, privacy and reproducibility concerns. To address these issues, this study presents the first systematic examination of the efficacy of both proprietary (GPT-3.5, and GPT-4) and open-source LLMs (LLAMA 7B,13B, and 70B) for the task of patient-trial matching. Employing a multifaceted evaluation framework, we conducted extensive automated and human-centric assessments coupled with a detailed error analysis for each model. To enhance the adaptability of open-source LLMs, we have created a specialized synthetic dataset utilizing GPT-4, enabling effective fine-tuning under constrained data conditions. Our findings reveal that open-source LLMs, when fine-tuned on this limited and synthetic dataset, demonstrate performance parity with their proprietary counterparts. This presents a massive opportunity for their deployment in real-world healthcare applications. To foster further research and applications in this field, we release both the annotated evaluation dataset along with the fine-tuned LLM -- Trial-LLAMA -- for public use.
\end{abstract}


\section{Introduction}\label{sec1}

Clinical trials represent both the most important and the most challenging aspect of medical advancements. These trials serve a dual function: first, as a conduit for patients to access potentially life-altering treatments, and second, as a mechanism for the iterative process of drug development and approval. However, a significant number of trials are beleaguered by extended timelines. Empirical data suggests that, on average, clinical trials take approximately twice as long as initially projected \cite{assessingstartup}, with approximately 40\% of trial sites failing to meet their enrollment targets \cite{enroll}. Apart from others, one of the major challenges in recruiting patients is matching them against suitable trials \cite{ bennette2016predicting, kadam2016challenges, berger2016opportunities, acrptufts2017, fayter2007systematic, automated, recruiting, effort}.

The process of matching a patient to trials is challenging. It requires both the meticulous analysis of electronic health records (EHRs) and the contextual interpretation of this data against the backdrop of clinical trial criteria. This is particularly challenging because a majority of this data is stored in unstructured documents written in free text. Even the structured data is difficult to query due to the increasing complexity of inclusion and exclusion criteria.

Automating this process can accelerate trials save healthcare providers' time spent on manual chart reviews. Current approaches primarily rely on data extraction or classification pipelines \cite{cohortselection, criteria2query, parker}. Nonetheless, these methods require extensive variable engineering, which frequently results in constrained contextual comprehension and limited scalability when dealing with intricate trial criteria.

The emergence of Large Language Models (LLMs), such as Med-PaLM \cite{Singhal2023} and GPT-4 \cite{Nori2023}, marks a paradigm shift in the domain of automated interpretation of patient health records. These models embody the cutting-edge in natural language processing (NLP), facilitating nuanced and context-aware analysis of complex medical data. Leveraging their capabilities, recent research has used these models for a variety of clinical information interpretation tasks, including patient matching \cite{trialgpt}. However, their deployment in healthcare settings presents challenges.

One primary concern relates to the risk of Protected Health Information (PHI) leakage when using such models. Most healthcare organisations prefer on-premise infrastructure for tools that handle identified patient data. However, due to the cost and computational complexity associated with these models, they often remain in centralized cloud provider environments. These challenges can make LLMs prohibitive for widespread clinical application. Moreover, despite their effectiveness, advanced models are often characterized by opacity and proprietary restrictions, which further complicate their integration into healthcare systems subject to stringent regulatory constraints.

In light of these considerations, there is a growing need for the development of open-source LLMs that can match the accuracy of their proprietary counterparts but at a significantly reduced cost. This also enables healthcare organizations to seamlessly integrate these technologies into their existing infrastructures, mitigating the risk of Protected Health Information (PHI) leaks. 

To the best of our knowledge, this study is the first comprehensive examination of the efficacy of open-source LLMs in this domain. Our contributions are as follows -
\begin{itemize}
    \item Our work thoroughly compares open-source LLMs and their proprietary counterparts for patient-trial matching.
    \item We further explore and elucidate the impact of fine-tuning on various open-source LLMs for patient-trial matching.
    \item We define the error taxonomy and thoroughly analyze the nature of errors made by the models on this task.
    \item Along with the experimental details, we publicly release the evaluation dataset and the LLM trained based on LLAMA for patient-trial matching.
\end{itemize}

\section{Related Work}
\subsection{Patient Matching} 

Leveraging unstructured patient records for the purpose of clinical trial recommendation has garnered significant attention in both academic and industrial realms. The process of patient-clinical trial matching can be bifurcated based on the search directionality into two distinct paradigms: the "trial-to-patient" approach \cite{cohortselection, criteria2query, parker, humanmachine}, and the "patient-to-trial" methodology \cite{trialgpt, trecdataset, trec2022, sigirdataset, reranking1, reranking2}. Some approaches even use a "patient-to-patient" matching strategy where known eligible patients are matched against unknown eligible patients \cite{casebased}. In the "trial-to-patient" paradigm, however, the process involves the ranking (or querying) of patients according to the inclusion and exclusion criteria of a specific trial. This approach usually entails extracting information from clinical trial criteria to make it structured \cite{criteria2query, elixr, electronicscreening} and then using the structured information to query the relational EHR database. Conversely, the "patient-to-trial" paradigm entails the identification of relevant trials for a patient, based on their EHRs. Our research is primarily concentrated on the latter paradigm. While there exists a substantial corpus of research utilizing neural ranking algorithms in this context \cite{reranking1, reranking2}, explorations into the use of generative LLMs, such as Generative Pre-trained Transformer (GPT) variants, have been relatively limited. Notably, these generative models have demonstrated superior performance over previous methodologies on almost all NLP tasks \cite{trialgpt}. Building upon the work presented in \cite{trialgpt}, our research extends this framework to incorporate and evaluate more open-source LLMs.

\subsection{Large Language Models}

LLMs such as GPT4 \cite{gpt4}, GPT3 \cite{gpt3}, PaLM \cite{palm}, and LLAMA \cite{llama2} have shown remarkable capabilities on different NLP tasks such as text summarization, information extraction and question answering. Large Language models have enormous potential to affect biomedical applications \cite{tian2023opportunities}. A significant body of research has been conducted to evaluate their performance in biomedical and clinical domain for summarization \cite{evidencesummarization, ctsummarization}, generation \cite{gpt4medicine}, interpretation\cite{infoextraction}, prediction \cite{gao2023leveraging} and information extraction \cite{llmextractors, alazraki2023not}. Despite their potential, there has been limited work on using their potential for patient-trial matching. Furthermore, instruction-tuning, which is a technique for training LLMs to perform specific tasks based on explicit instructions, has shown to improve their performance by a significant margin \cite{instructiontuning, mishra2022cross, instructiontuning3}. However, there is a lack of instruction-tuning dataset for healthcare despite several attempts at building it \cite{ctsummarization, medalign, zhou2023survey}. Meanwhile, distillation has proven to be very effective in improving the performance of LLMs when there is lack of human curated dataset \cite{gpt4tuning, alpaca, vicuna, zephyr, self}. In this study, we leverage instruction-tuning and distillation to improve the performance of different open-source LLMs on the task of patient-trial matching.

\section{Methodology}

\begin{figure}[h]
\centering
\includegraphics[width=\linewidth]{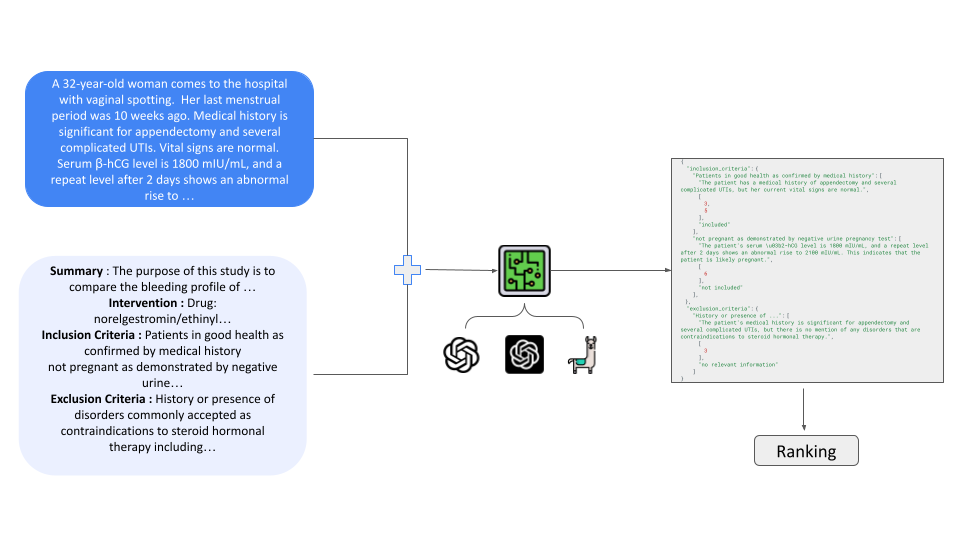}
\caption{Illustration of the Output Pipeline for Trial Ranking. The input and output JSON have been shortened for brevity. The model receives both patient summaries and inclusion and exclusion criteria of clinical trial. Then such information was transferred to JSON format, which contains the explanation for each criteria, reference sentences, and final predicted labels. }
\label{fig:output_pipeline}
\end{figure}
Our pipeline for ranking clinical trials builds upon the previous work outlined in \cite{trialgpt}. We don't make any significant changes to their pipeline. We, however, explain the overall pipeline briefly here as well. The process initiates by generating criterion-level explanations for each trial, pertinent to the patient's summary, utilizing a LLM as the backbone. This procedure is graphically represented in Figure \ref{fig:output_pipeline}. Specifically, for each criterion and corresponding patient summary, the LLM performs the following tasks:
\begin{enumerate}
    \item[(a)] Produces a step-by-step explanation employing a Chain of Thought reasoning (CoT) approach.
    \item[(b)] Identifies and utilizes specific sentences from the patient note to underpin its reasoning.
    \item[(c)] Arrives at a final decision, categorizing the patient as either \texttt{included} or \texttt{not included} for inclusion criteria, and \texttt{excluded} or \texttt{not excluded} for exclusion criteria. In scenarios where the patient note lacks relevant information, the model outputs \texttt{no relevant information.}
\end{enumerate}

This model output is aligned to the structured format through the use of carefully crafted prompts, detailed in the Appendix. Subsequently, the rank score for a trial with regards to a patient is computed as the difference between the percentage of met inclusion criteria and the percentage of met exclusion criteria for that patient-trial combination following \cite{trialgpt}.

\begin{equation}
    R = \% \text{ met incl. criteria} - \% \text{ met excl. criteria}
\end{equation}

where R is the rank score.
While \cite{trialgpt} use additional model-generated score as well, we don't include that score for simplicity. Similarly, the exclusion of a clinical trial is determined using the formula:

\begin{equation}
    E = I(\text{\% unmet incl. criteria} > 0) + I(\text{\% met excl. criteria} > 0) - \text{\% met incl. criteria}
\end{equation}

where $I$ is an indicator function and $E$ is the exclusion score. 

\subsection{LLMs and Experimental Setup}

 We tested both proprietary (GPT-3.5, and GPT-4) and open-source LLMs (LLAMA-2 7B,13B, and 70B, referred to LLAMA hereafter) for the task of patient-trial matching. For GPT-3.5, we leveraged the Azure Open AI API, specifically \texttt{gpt-35-turbo-16k-0613} as the model version. We set the temperature parameter to 0, aiming for deterministic outputs that would ensure consistency and repeatability in our experiments. This was coupled with a \texttt{top\_p} setting of 0.95, aligning with our goal to eliminate randomness in the model's response generation process. Additionally, we refrained from applying any frequency or repetition penalties, allowing the model's natural language generation capabilities to function without these constraints. For GPT-4, we employed a similar configuration with \texttt{gpt-4-0613} as the model version. For LLAMA, we changed the configuration from Meta. We initially encountered challenges in aligning the standard versions of these models to produce outputs in the required format, particularly in the context of complex clinical trial criteria. To address this, we opted for specific versions tailored for chat applications, namely \texttt{Llama-2-7b-chat-hf}, \texttt{Llama-2-13b-chat-hf}, and \texttt{Llama-2-70b-chat-hf}. These versions offered a more flexible and adaptable framework for our needs. We adjusted the temperature setting to 0.4 for all LLAMA models, a decision informed by preliminary tests which indicated that a slightly higher temperature prevented the model from collapsing on certain trials where inclusion/exclusion criteria were not clearly defined. Maintaining the output format was particularly challenging when working with the base LLAMA models. Despite employing various techniques such as context-free grammar (CFG) to constrain the model's output, the results remained suboptimal. Consequently, the models were unable to generate structured output for complex clinical trials. To circumvent this, we adjusted the model's temperature to foster more exploratory behavior and executed five output generations iteratively till the output matched our JSON schema. This allowed us to generated structured output for majority of clinical trials even with base models.

\subsection{Fine-tuning}
\begin{figure}[h]
\centering
\includegraphics[width=\linewidth]{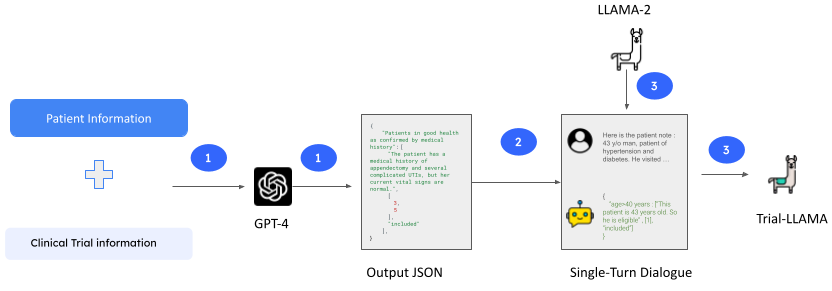}
\caption{Illustration of the fine-tuning for LLAMA models. (1) GPT-4 generates output for the sampled clinical trial-patient pairs (2) Its output is transformed into a single-turn dialogue generation task (3) LLAMA is trained on the dialogue generation tasks using SFT }
\label{fig:finetuning}
\end{figure}

For the open-source LLMs, we fine-tuned all three variants of LLAMA on the overall generation process, including explanation of each clinical trial criterion, evidence in patient summaries, and label prediction, rather than limiting to the final label prediction. Figure \ref{fig:finetuning} illustrates the fine-tuning process. We first generated the output for 2,000 patient-trial pairs sampled from the training set using GPT-4 using the pipeline discussed in Figure \ref{fig:output_pipeline}. This output format was chosen to capture a comprehensive view of the model's reasoning process, including its ability to reference evidence and its final decision-making. Then we used the dataset of patient-trial-output triplet, as generated using GPT-4 to fine-tune LLAMA models. We utilized standard supervised fine-tuning loss as our primary optimization criterion over these outputs. Unlike some methods that involve parameter reduction techniques like quantization or low-rank adaptations, our approach involved full-scale fine-tuning.  The patient-trial combination was given as input to the model as prompt and then output was used as to train the model using standard cross entropy loss for fine-tuning of the model. We also added our instruction(as explained in Appendix) as system prompt for all training inputs.

We also restructured the input and output format for the the patient-trial pair dataset into a system-user-assistant framework. This structure was adopted to mitigate catastrophic forgetting of LLMA, as it had been specifically tailored for dialogue generation using the same format. In our fine-tuned LLAMA models, we chose to forego the use of exemplars in prompts during the inference phase. This strategic decision was grounded in our empirical observations, which indicated that the incorporation of exemplars did not yield a substantial improvement in performance. However, it did lead to a reduction in the usable context size window of the model, preventing it from processing longer inputs.
We conducted the fine-tuning experiments on a machine with 16 Nvidia A100 GPUs, each equipped with 80GB of GPU RAM. The collective duration of model training amounted to approximately 2-3 hours.

Please note that, in our study, we adhered to fine-tuning only the LLAMA model variants. It's important to note that while the option to fine-tune GPT-3.5 was available (fine-tuning GPT-4 was not available to public at the time of writing this paper), we consciously chose not to employ fine-tuning for this model. The primary goal of our study was not to establish the superiority of open-source models over proprietary ones like GPT-3.5 in general terms. Instead, our focus was on refining these open-source models so that they could meet or exceed the benchmarks set by proprietary models like GPT-3.5. This way, they offer viable, cost-effective alternatives in healthcare applications where using proprietary models might be prohibitive due to privacy or other constraints. 

\subsection{Evaluation}

In line with the methodologies adopted in \cite{trialgpt}, our evaluation is conducted at both aggregate-level and criterion-specific level, allowing for a comprehensive assessment of model performance.

In the aggregate-level analysis, the model's utility is quantified by its ability to accurately rank clinical trials. This is achieved by first leveraging criterion-level outputs as a basis for trial ranking. To rigorously evaluate this ranking, we employ two key metrics: Normalized Discounted Cumulative Gain at the 10th item (NDCG@10) and Precision at the 10th item (Precision@10). Additionally, we incorporate the Area Under the Receiver Operating Characteristic (AUROC) curve as a statistical measure to discern the model's accuracy in differentiating between patient inclusion and exclusion in a trial. This dual approach, encompassing both ranking quality and classification accuracy, provides a comprehensive view of the model's performance at the aggregate level.

For criterion-level performance, we manually annotated patient summary-criteria combination. We first generated the results for test set and flattened the output in the form of (patient, trial, criteria, model-output). Thus, each row in the flattened output consisted of the out of the model for specific criteria -- patient combination. A corpus of 500 criteria was then selected and labeled across five categories: 'included', 'not included', 'excluded', 'not excluded', and 'no relevant information,' depending on their designation as inclusion or exclusion criteria.

The selection of criteria to be annotated was based on whether that specific criteria is diverse and complex enough. Instead of sampling random inclusion/exclusion criteria and annotating them, we developed the algorithm shown in Algorithm \ref{alg:novel_criterion_selection}. We first generate initial predictions using GPT-4, and form a preliminary set denoted as \(\mathcal{C}_{\text{pred}}\). We use all the test examples to generate these outputs. Criteria mentioning 'age' or 'gender' are excluded, yielding a reduced set, denoted as \(\mathcal{C}_{\text{reduced}}\). From \(\mathcal{C}_{\text{reduced}}\), we select a subset \(\mathcal{C}_{\text{selected}}\) of approximately 500 criteria for each label (total 2500 outputs), based on GPT-4 predictions. We acknowledge that GPT-4 does not reflect accurate outputs, this method makes it easier to perform the selection with minimal compromise on the diversity of labels. We further filter \(\mathcal{C}_{\text{selected}}\) to create a subset of criteria. This step is based on the algorithm used by \cite{self} for selection of novel tasks to train GPT-3.5 using distillation. We then generate \(\mathcal{C}_{\text{final}}\) by sampling 100 criterion per label from \(\mathcal{C}_{\text{novel}}\). Each criterion in \(\mathcal{C}_{\text{final}}\) is further classified as 'implicit' or 'explicit', leading to two distinct accuracy measures: Explicit Criterion Level Accuracy (CLA) and Implicit CLA. 

\begin{algorithm}
\caption{Novel Criterion Selection Based on ROUGE Score}
\label{alg:novel_criterion_selection}
\begin{algorithmic}[1]
\Require $ \mathcal{C}_{\text{selected}} $: A set of 2500 criteria, 500 for each label
\Ensure $ \mathcal{C}_{\text{novel}} $: A set of novel criteria

\State Initialize an empty set $ \mathcal{C}_{\text{novel}} $

\For{each label $ l $ in criterion labels}
    \For{each criterion $ c $ in $ \mathcal{C}_{\text{selected}} $ corresponding to label $ l $}
        \State Compute ROUGE Score: $ \text{score} \leftarrow \text{ROUGE}(c, \mathcal{C}_{\text{novel}}) $
        \State Compute Novelty Index: $ \text{noveltyIndex} \leftarrow 1 - \frac{\text{score}}{\tau} $
        \If{$ \text{noveltyIndex} > 0 $}
            \State Add $ c $ to $ \mathcal{C}_{\text{novel}} $
        \EndIf
    \EndFor
\EndFor

\State \Return $ \mathcal{C}_{\text{novel}} $
\end{algorithmic}
\end{algorithm}


Each criterion in \(\mathcal{C}_{\text{final}}\) is then annotated with a gold-standard answer and corresponding evidences. These evidential references serve as a basis for language models to substantiate their answers. To gauge the faithfulness of various models in accurately citing these pieces of evidence, we calculate precision, recall, and F1 scores for each model. Additionally, we conduct a direct comparison of model performance at the criterion level to evaluate their relative effectiveness.

Different from the metrics used in \cite{trialgpt}, we created two distinct aspects of Criterion-Level Accuracy (CLA), namely Explicit CLA and Implicit CLA, to holistically assess the model's performance. For Explicit CLA, our focus is on evaluating how accurately the model categorizes each criterion into the correct class, provided that the criterion has been previously identified as 'explicit' in our manual annotation exercise. This evaluation primarily concerns criteria for which the necessary information for classification is clearly and directly stated in the patient documentation, leaving minimal room for interpretation or inference. The accuracy here reflects the model's ability to comprehend and correctly apply these straightforward, unambiguous data points.

On the other hand, Implicit CLA tackles a more nuanced challenge: it assesses the model's performance on criteria deemed 'implicit' by the annotators. These criteria involve situations where the required information is not explicitly stated but rather implied or inferred from the available data. This often requires connecting disparate pieces of information, understanding subtleties and nuances in the patient data, and making educated guesses based on the context. Calculating the Implicit CLA involves a thorough analysis of how well the model navigates these complexities and accurately classifies criteria based on less direct information.

Both Explicit and Implicit CLAs are pivotal in understanding the model's overall capability to process and interpret clinical trial criteria. While Explicit CLA provides insight into the model's proficiency with clear-cut, straightforward tasks, Implicit CLA sheds light on its ability to handle ambiguity and complexity – crucial aspects in the realm of clinical data interpretation.

\subsection{Dataset}

For our dataset, we adopted the similar datasets as used in \cite{trialgpt} that incorporate the SIGIR dataset \cite{sigirdataset} and both the 2021 and 2022 versions of the TREC CT cohorts \cite{trec2022, trecdataset}, as shown in Table \ref{tab:dataset_stats}. For each patient within these datasets, we extract 50 clinical trials, categorizing them into three distinct classifications: "eligible", "excluded", and "irrelevant".

\begin{table}[h]
\centering

\setlength{\tabcolsep}{3.5pt} 
\begin{tabular}{|l|c|c|c|}
\hline
\textbf{Metric} & \textbf{Test Set} & \textbf{Train Set*} & \textbf{Sample Train Set} \\ \hline
Patient-Trial Pairs & 4678 & 22564 & 2000 \\ \hline
\# of Patients & 36 & 140 & 137 \\ \hline
\# of Trials & 4452 & 18828 & 1930 \\ \hline
Total Trials/Patient & 129.944 ± 100.130 & 161.171 ± 104.507 & 14.599 ± 9.838 \\ \hline
\# of irrelevant trials/patient & 45.667 ± 8.966 & 47.629 ± 7.236 & 4.438 ± 2.007 \\ \hline
\# of ineligible trials/patient & 39.972 ± 58.452 & 54.4 ± 62.809 & 5.0 ± 6.289 \\ \hline
\# of eligible trials/patient & 44.306 ± 48.046 & 59.143 ± 58.883 & 5.161 ± 5.451 \\ \hline
\# of words/ patient & 95.194 ± 32.175 & 110.729 ± 40.901 & 111.423 ± 40.976 \\ \hline
\# of sentences/patient & 7.5 ± 3.403 & 8.579 ± 3.402 & 8.613 ± 3.403 \\ \hline
\end{tabular}
\caption{Statistics of the training, test and the sampled training set. Please note that only 2000 patient-trial pairs are sampled from the training set for the final training}
\label{tab:dataset_stats}
\end{table}

The categorization within the SIGIR dataset required a different approach, given its classification system. The SIGIR cohort's classes are as follows:
\begin{enumerate}
    \item[(a)] "Will not refer to the trial": This class aligns with the 'irrelevant' category in our study.
    \item[(b)] "Will refer to the trial": Corresponds to the 'eligible' category.
    \item[(c)] "May refer to the trial": This class does not map directly to any of our predefined categories.
\end{enumerate}
Due to this non-conformity, we excluded all trial-patient combinations classified under (c) "May refer to the trial", to maintain consistency.

To facilitate the fine-tuning of our models, we partition the dataset into a training and test set, adhering to an 80:20 ratio. This division is implemented along the patient axis to ensure no test patient record gets leaked into the training set. Prior to splitting, all datasets are combined and thoroughly shuffled. The specifics of the training and test sets are displayed in Table \ref{tab:dataset_stats}. It is noteworthy that despite the large volume of records in the training set, they are not fully utilized for model training. Instead, the large size of this set provides with an easy mechanism to sample diverse training samples for fine-tuning while also allowing us to save on compute costs associated with evaluation a large number of model checkpoints. Evidently, as shown in Table \ref{tab:dataset_stats} the sampled dataset is more diverse than the training set.

\section{Results}\label{sec2}

\subsection{Overall Comparison}
As mentioned in the previous section, the performance evaluation of these models is conducted across multiple dimensions, encompassing:
\begin{enumerate}
    \item[(a)] The model's proficiency in accurately determining whether a patient satisfies a given criterion (Explicit CLA, and Implicit CLA)
    \item[(b)] The ability of the model to cogently and coherently explain its reasoning process while citing correct evidence (Evidence faithfulness )
    \item[(c)] The effectiveness of the criterion level accuracy in ranking clinical trials for a patient (NDCG@10, P@10, AUROC)
\end{enumerate}

Table \ref{tab:performance} shows the performance of different LLMs for trail ranking and criterion-level evaluation. Notably, the fine-tuned LLAMA 70B model, referred to as Trial-LLAMA 70B, demonstrates superior performance compared to all publicly available open-source alternatives, and it even competes favorably with, and in certain instances surpasses, GPT-3.5. Remarkably, it achieves these results despite having been trained solely on synthetic data.

\begin{table}[ht]
  \centering
  \setlength{\tabcolsep}{3.5pt}
  \begin{tabular}{cccccc} 
    \hline
    \textbf{Model} & \textbf{NDCG@10} & \textbf{Precision@10} & \textbf{AUROC} & \textbf{Implicit CLA }& \textbf{Explicit CLA} \\
    \hline
    GPT3.5 & 67.57         & 60.34              & 65.69          & 56.85                 & 54.2 \\ 
    GPT4 & 77.28         & 70.05              & 73.90          & 75.31                 & 58.8 \\
    LLAMA 7B           & 41.47         & 29.9               & 56.8           & 26.59                 & 28.53\\
    LLAMA 13B & 45.66         & 39.59              & 54.37          & 39.07                 & 53.3  \\
    LLAMA 70B &43.58         & 39.45             & 54.37          & 29.24                 & 45.41 \\
    \hline
    \textbf{Trial-LLAMA 70B} &\textbf{66.36 }        & \textbf{58.86  }            & \textbf{65.28  }        & \textbf{68.77}                 & \textbf{59.9} \\ 
    \hline 
  \end{tabular}
  \caption{Table showing the performance of different models on ranking as well as criterion level metrics}
  \label{tab:performance}
\end{table}

\subsection{Comparative Superiority of GPT-4 Over GPT-3.5}
\begin{figure}[ht]
  \centering
  \begin{subfigure}[t]{0.45\linewidth}
    \includegraphics[width=\linewidth]{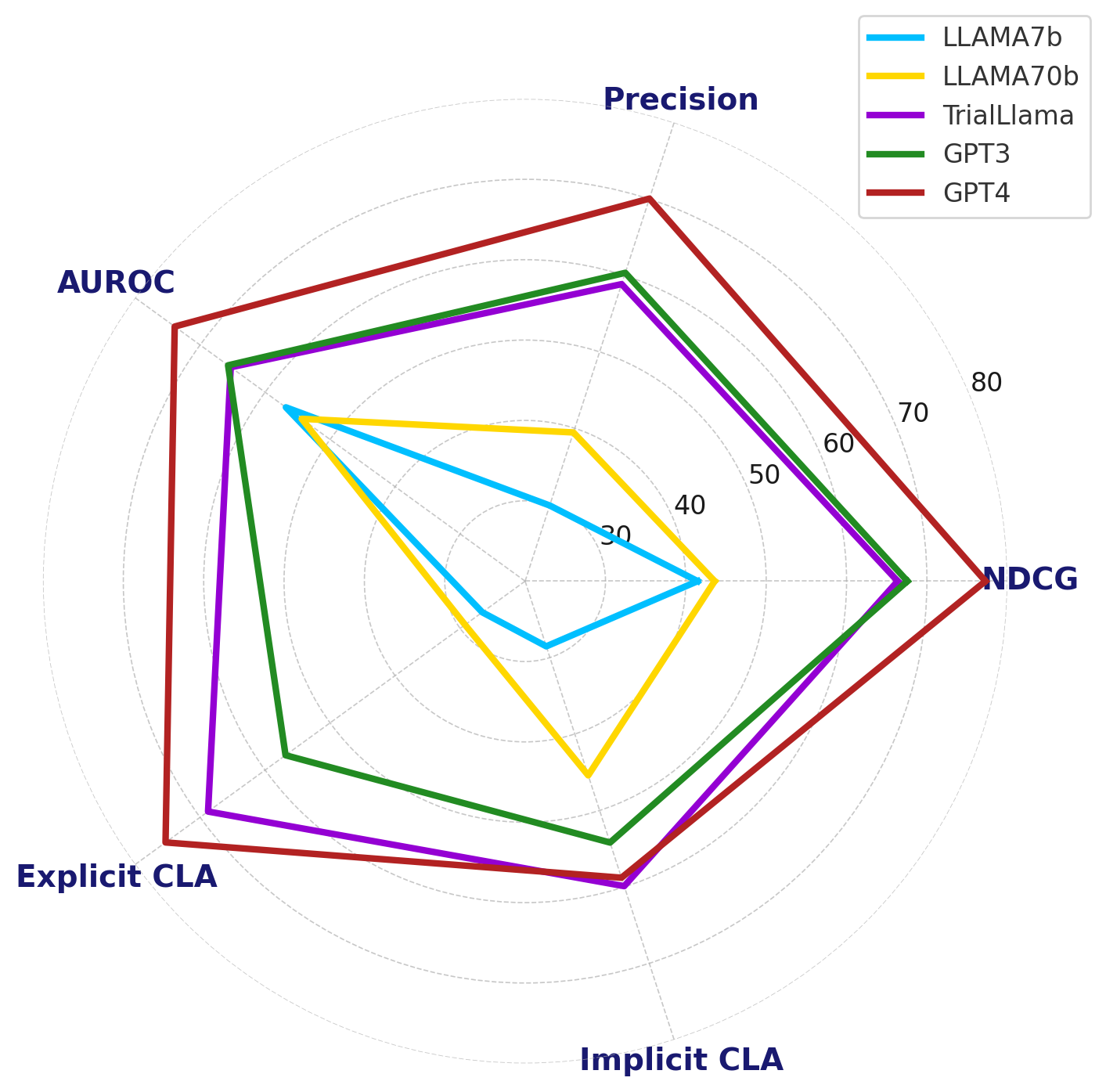}
    \caption{Spiderchart comparing the performance of different language models, including their fine-tuned variants on different metrics}
    \label{fig:spiderchart}
  \end{subfigure}
  \hfill
  \begin{subfigure}[t]{0.47\linewidth}
    \centering
    \includegraphics[width=\linewidth]{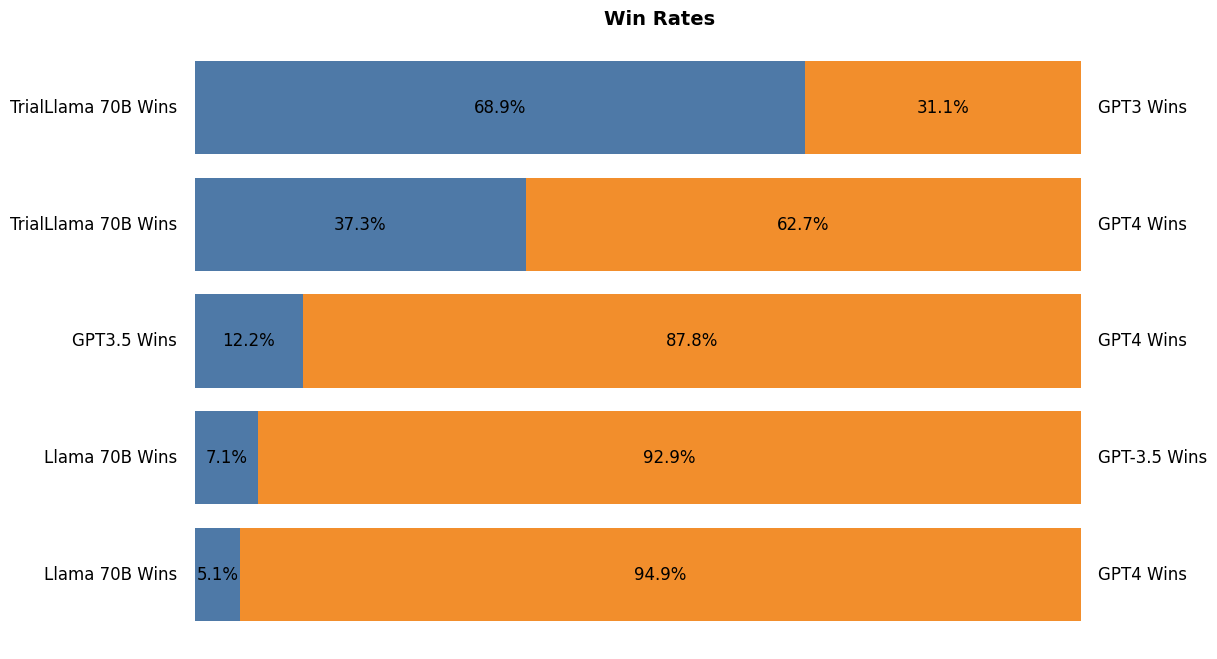}
    \caption{Head to Head comparison of different models on criterion level answers where there is disagreement between the models}
    \label{fig:winrate}
  \end{subfigure}
  \caption{Charts showing the overall performance of proprietary, open source as well as our proposed fine-tuned variant on different metrics along with their head-to-head comparison on each criterion where they disagree with each other}
  \label{fig:overall}
\end{figure}

Our empirical results unequivocally demonstrate that GPT-4 clearly outperforms GPT-3.5, as illustrated in Figure \ref{fig:spiderchart} and \ref{fig:winrate}. This superiority is not merely marginal but is pronounced across all five evaluation metrics. Specifically, GPT-4 exhibits enhanced precision in its outputs and demonstrates a more refined capability in ranking tasks. Furthermore, a noteworthy aspect of GPT-4's progress lies in its implicit reasoning ability. Our data indicates that GPT-4's performance in tasks requiring implicit reasoning surpasses that of GPT-3.5 by a considerable margin. Even after subjecting various models to rigorous fine-tuning procedures, none achieved the comprehensive proficiency exhibited by GPT-4. Although the most advanced iteration of our fine-tuned model showed some improvement in specific metrics, it still fell short of GPT-4's overall excellence. This discrepancy highlights GPT-4's advanced architectural and algorithmic enhancements, which contribute to its superior performance.

\subsection{Importance of Alignment for Specialized Tasks}

While the size and architectural design of a model is undoubtedly influential, we contend that alignment (which can be defined as the ability of the language model to follow the prompt instructions as closely as possible) is equally, if not more, crucial for success, especially in specialized applications like patient-matching. The empirical data presented in Table \ref{tab:performance} substantiates this assertion.

A notable example is the comparison between the LLAMA-13B and LLAMA-70B models. Contrary to expectations based solely on model size, LLAMA-13B demonstrates superior performance over its larger version, LLAMA-70B, in terms of all metrics. This observation suggests that the difference in performance is not inherently tied to the model's size but rather to its alignment with the specific task requirements. Our hypothesis is that both models encounter difficulties when attempting the task, and despite having greater computational power, LLAMA-70B fails to utilize it effectively, resulting in underperformance.

Further corroboration comes from the enhanced performance of the fine-tuned LLAMA-70B model, i.e., Trial-LLAMA-70B. Post fine-tuning, Trial-LLAMA-70B surpasses the performance of GPT-3.5, a model with approximately 175 billion parameters—more than double that of Trial-LLAMA. This underscores the significance of aligning a model with a specific task. It suggests that when a model is properly aligned and fine-tuned to the nuances of a particular task, its efficiency and effectiveness can significantly exceed those of larger models of larger models that are less aligned with that task. This principle holds particularly true for specialized tasks in healthcare like patient-trial matching, where there is limited publicly available labeled data. Even the Reinforcement learning from human feedback (RLHF) process, which has been employed to enhance these models' performance, might fail to incorporate a deep understanding of such nuanced tasks into the model.

\subsection{Challenges in Implicit Criterion Inference}

The inherent complexity of implicit criterion inference is evident from the data illustrated in Figure \ref{fig:spiderchart} and Table \ref{tab:performance}. Across all models, regardless of their size or the extent of fine-tuning, we observe a consistent trend that the performance on implicit criteria lags significantly behind that on explicit criteria. This finding aligns with our expectations since the task of semantic matching, which is predominant in explicit criteria, is intrinsically less complex than the reasoning required by implicit criteria. 

Despite this overall trend, it is noteworthy that fine-tuning yields substantial improvements in model performance for both types of criteria. This enhancement is clearly evident in the case of Trial-LLAMA, which, upon fine-tuning, outperforms GPT-3.5 in handling both implicit and explicit criteria. The disparity in performance between implicit and explicit criteria underscores the challenges posed by tasks that require deeper levels of reasoning and understanding. While semantic matching in explicit criteria relies more on surface-level text matching, implicit criteria demand a more profound comprehension and inferential reasoning. This is the potential reason why even advanced LLMs achieved only moderate performance.

\subsection{Head-to-Head Comparison of LLMs} 

For head-to-head analysis, we compare the win rates of a LLM against another LLM. If the responses of two models for the same criterion are different from each other, we check which of the two models wins in that scenario. Figure \ref{fig:winrate} shows the head-to-head analysis comparison results. The comparative analysis of Trial-LLAMA 70B and GPT-3.5 reveals an intriguing finding. On an aggregated ranking level, Trial-LLAMA 70B's performance is comparable to that of GPT3.5. However, a deeper dive into criterion-level results unveils a different narrative. As illustrated in Figure \ref{fig:winrate}, Trial-LLAMA 70B holds a significant edge over GPT-3.5 in head-to-head comparisons. Interestingly, Trial-LLAMA 70B also demonstrates commendable competitiveness against the GPT-4 variant, indicating its robustness in nuanced language tasks.

We posit that this discrepancy can be attributed to the intricacies of our overall pipeline, as depicted in Figure \ref{fig:output_pipeline}. Notably, the model's role extends beyond generating outputs for each criterion; it also \textit{decides} which criteria to generate outputs for. This decision-making process contributes to a disparity in criteria selection across models. For instance, we observed approximately an 80\% overlap (measured using a ROUGE score threshold of 0.9) between the criteria selected (and analyzed) by GPT-3.5 and GPT-4 while generating the output from the same clinical trial information. Similar trends are noted with other models. This indicates that the model separately addresses the inclusion and exclusion criteria of the trial while generating output, which can have a measurable impact on the final ranking using our formulas.

Therefore, while fine-tuning leads to improvements in individual criterion performance, this enhancement does not proportionately translate into the final aggregate ranking metric, as it also changes the way the model handles the inclusion and exclusion criterion. Consequently, to accurately gauge the models' understanding and performance at the criterion level, a head-to-head comparison on the \textit{same} criteria, rather than solely relying on overall ranking metrics, is more robust, illustrative, and informative.

Figure \ref{fig:winrate} also highlights a substantial performance gap between the non-fine-tuned LLAMA version and GPT-3.5, further underscoring the impact of fine-tuning on model efficacy.

\subsection{Evidence Faithfulness in LLM Outputs}
\begin{figure}[ht]
    \centering
    \includegraphics[width=0.85\linewidth]{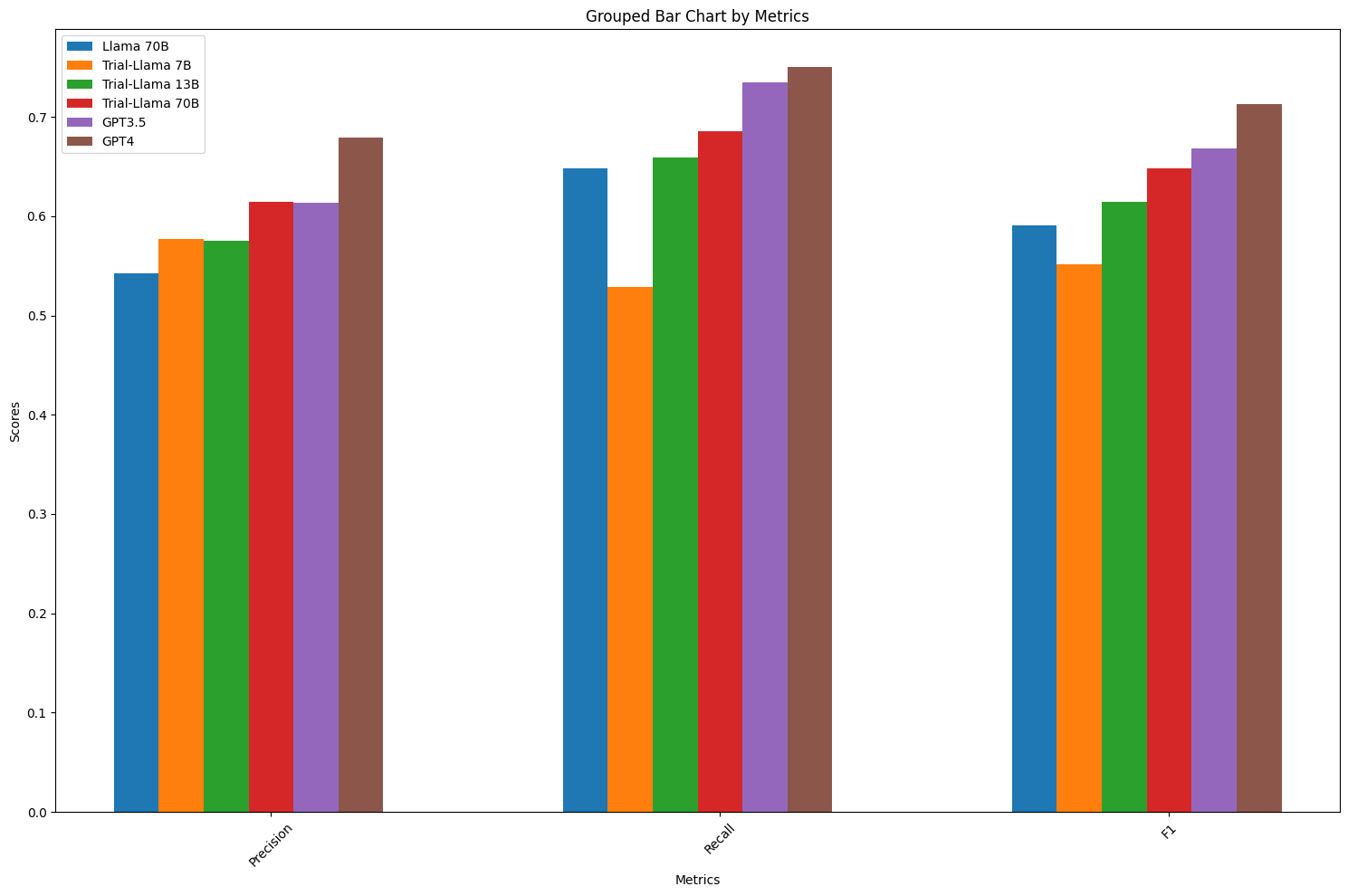}
    \caption{Performance of different models in referencing correct evidence while generating the output at criterion level}
    \label{fig:evidence}
\end{figure}

In addition to assessing the accuracy of models in correctly classifying each criterion, the fidelity of the models in referencing the correct pieces of evidence is of significant importance. This aspect of evidence faithfulness is critical, particularly when considering the deployment of such systems in real-world scenarios. It is essential to instill confidence in end users, often clinicians or researchers, regarding the reliability of the model's outputs. LLMs are known to occasionally produce outputs based on internal heuristics or 'hallucinations' rather than concrete evidence. The ability to anchor these models' responses to explicit, original evidences significantly enhances the trustworthiness of the system. It allows end users to discern whether the model's outputs are based on factual data or are a result of erroneous generation processes.

To rigorously evaluate this aspect of model performance, we manually annotated sentences within patient notes that are pertinent to specific criteria. This annotation process establishes a ground truth against which the model's reference to evidence can be evaluated. We then computed and compared the precision and recall of various models in referencing these annotated evidences. As depicted in Figure \ref{fig:evidence}, the trends in evidence faithfulness align with those observed in accuracy evaluations. This correlation suggests that models with higher accuracy in criterion classification also tend to be more reliable in citing appropriate evidence, implying that improvements in model accuracy do not occur in isolation but are accompanied by enhancements in the model's ability to reference relevant and accurate information.

\subsection{Ablation Study Results of Fine-tuning}

\begin{figure}[ht]
  \centering
  \begin{subfigure}[t]{0.45\linewidth}
    \includegraphics[width=\linewidth]{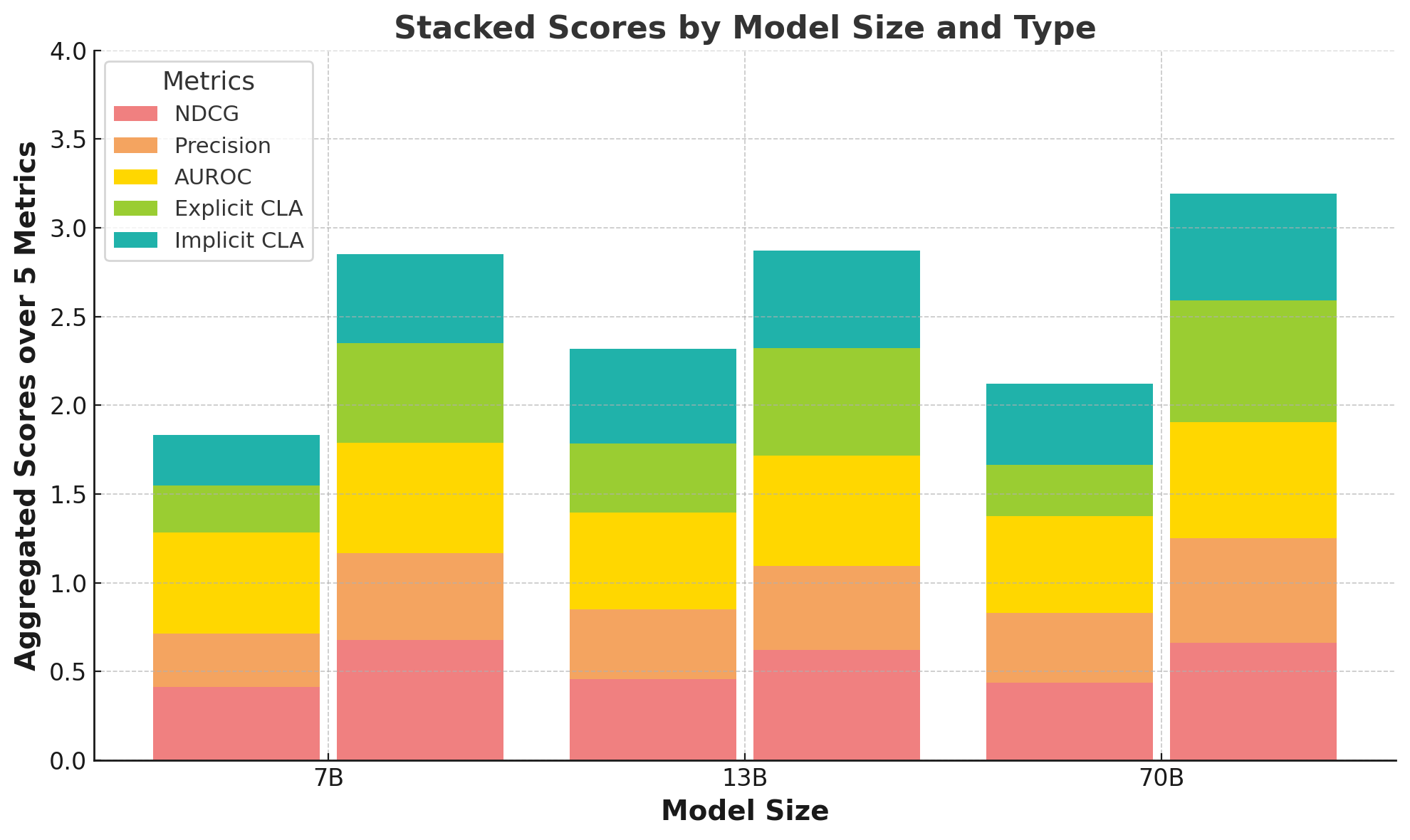}
    \caption{Comparison of the aggregated and normalized performance of the models of different sizes with and without finetuning }
    \label{fig:size}
  \end{subfigure}
  \hfill
  \begin{subfigure}[t]{0.47\linewidth}
    \centering
    \includegraphics[width=\linewidth]{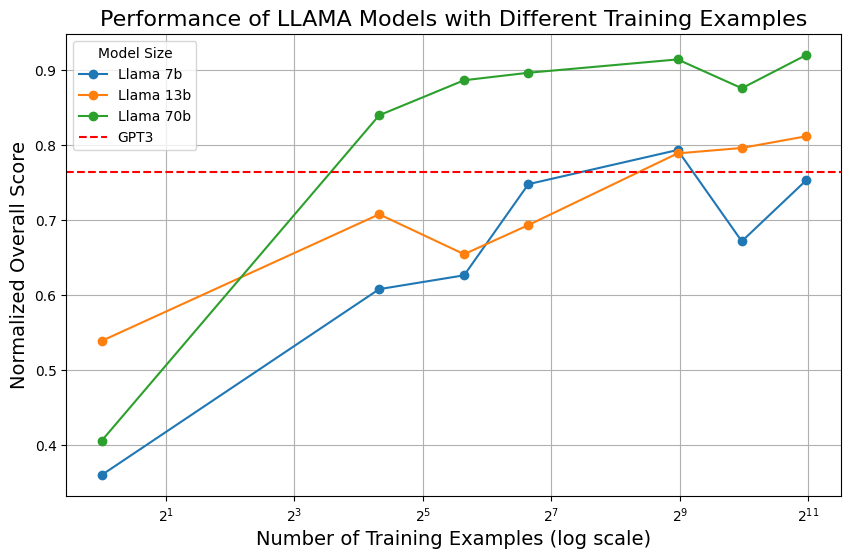}
    \caption{Variation of the performance of different sized models (normalized via GPT-4 score) with the increase in the number of fine-tuning examples (log scale with base 2)}
    \label{fig:data}
  \end{subfigure}
  \caption{Charts showing the impact of both model size as well as the size of the fine-tuning dataset on LLAMA variants }
  \label{fig:variation}
\end{figure}

\subsubsection{Performance Variation with Size} 

Without fine-tuning, the LLAMA 13B parameter model performs on par with the 70B parameter model and strongly outperforms the 7B parameter model (Figure \ref{fig:size}. However, following fine-tuning, the performance of LLAMA 70B model vastly improves over both the 13B  and 7B variants across all metrics. This observation suggests that while fine-tuning is beneficial, it continues to be influenced by the underlying pretraining knowledge and inherent capabilities of the model. Interestingly, the LLAMA 7B and 13B models exhibit similar performance levels post-fine-tuning, even though the 13B model is slightly superior. This is not unexpected, as the leap in parameters from 13B to 70B ) has a more significant impact on performance enhancement than the increase from 7B to 13B.

\subsubsection{Performance Variation with Data}

It is known that the performance of a model improves with the volume of data it is exposed to \cite{CHINCHILLA}. Nevertheless, the quality of data plays a pivotal role in determining the output's caliber. Multiple research works have demonstrated that while the fine-tuning performance of a model initially improves rapidly, it tends to reach a saturation point beyond a certain threshold of data exposure \cite{SOMEREFERENCE}. This phenomenon is consistent with our findings with the fine-tuning of the LLAMA models of different sizes. As illustrated in Figure \ref{fig:data}, the performance of all three LLAMA variants exhibits a significant initial leap with exposure to a small data subset, followed by a gradual enhancement as they are introduced to an increasing number of examples. Notably, the largest LLAMA variant swiftly surpasses the performance of GPT-3.5. For the assessment of model performance, we utilize the metric of overall criteria level accuracy, encompassing both implicit and explicit criteria.

\subsection{Error Analysis}
\begin{table}[h]
\centering
\begin{tabular}{|p{3cm}|p{5cm}|p{4cm}|}
\hline
\textbf{Error Type}              & \textbf{Description}                                                                                           & \textbf{Example}                                                                                     \\ \hline
Implicit Failure                 & Cases where the model fails to use implicit reasoning, leading to errors despite derivable information.        & Failure to infer a patient's condition from implicit data in their history.                          \\ \hline
Lack of Information              & Errors from assumptions based on absent explicit information, often in comorbidity contexts.                   & Assuming a breast cancer patient doesn't have lung cancer due to no explicit mention.                \\ \hline
Wrong Outcome                    & Direct misjudgments where the model's output is plainly incorrect.                                             & Misclassifying a patient with MDD as 'no relevant information' for a criterion needing a healthy patient. \\ \hline
Explanation-Output Mismatch      & Instances of correct explanations but incorrect final outputs, showing decision-making inconsistencies.        & Accurate reasoning but incorrect eligibility decision.                                               \\ \hline
Expert Opinion Needed            & Cases needing expert interpretation, but the model provides its own answer.                                    & Deciding on medication use requiring an investigator's discretion.                                   \\ \hline
Negated Criteria                 & Misinterpretation of negated criteria, especially in inclusion scenarios.                                      & Incorrectly assessing eligibility based on a negated medication criterion.                           \\ \hline
\end{tabular}
\caption{Classification and Examples of Error Types in Model Output}
\label{tab:error_types_compact}
\end{table}

\begin{figure}
    \centering
    \begin{subfigure}{0.33\textwidth}
        \includegraphics[width=\linewidth]{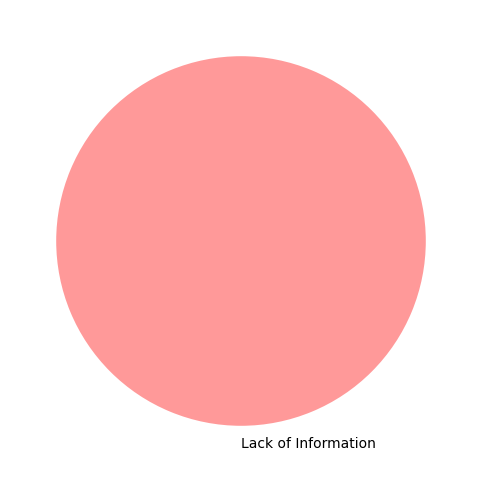}
        \caption{Errors for Criteria Labelled as 'Included'}
        \label{fig:included}
    \end{subfigure}\hfill
    \begin{subfigure}{0.45\textwidth}
        \includegraphics[width=\linewidth]{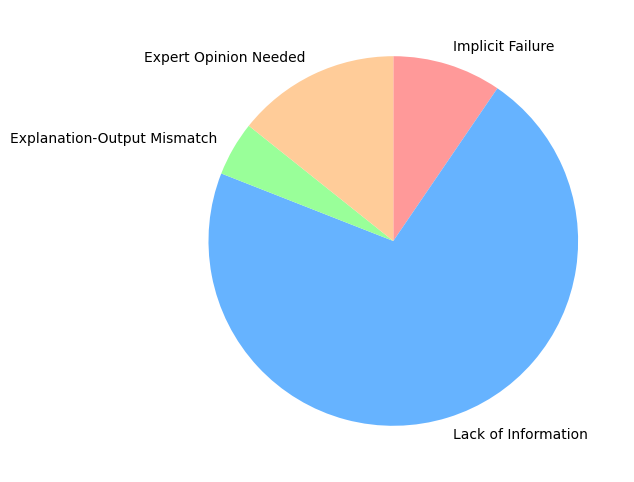}
        \caption{Errors for Criteria Labelled as 'Excluded'}
        \label{fig:excluded}
    \end{subfigure}

    \vspace{1cm} 

    \begin{subfigure}{0.36\textwidth}
        \includegraphics[width=\linewidth]{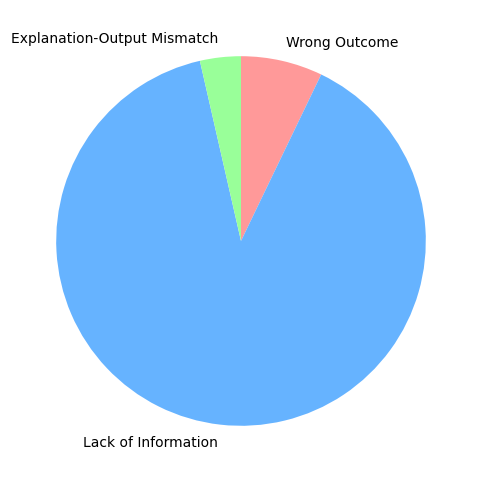}
        \caption{Errors for Criteria Labelled as 'Not Included'}
        \label{fig:notincluded}
    \end{subfigure}\hfill
    \begin{subfigure}{0.43\textwidth}
        \includegraphics[width=\linewidth]{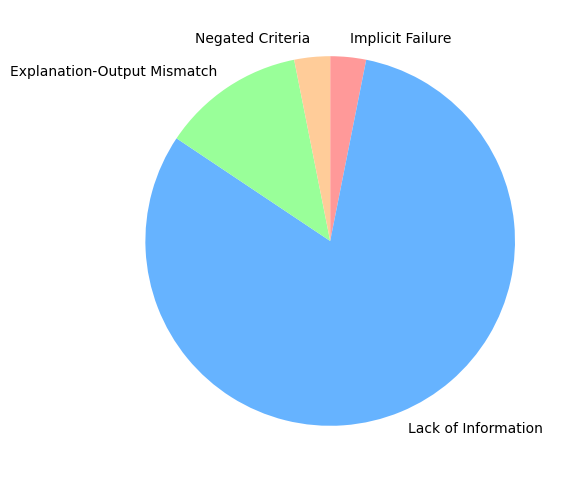}
        \caption{Errors for Criteria Labelled as 'Not Excluded'}
        \label{fig:notexcluded}
    \end{subfigure}

    \vspace{1cm} 

    \begin{subfigure}{\textwidth}
        \centering
        \includegraphics[width=0.38\linewidth]{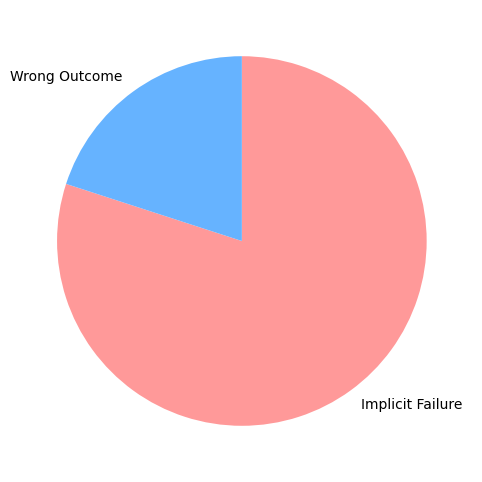}
        \caption{Errors for Criteria Labelled as 'No Relevant Information'}
        \label{fig:norelevant}
    \end{subfigure}
\caption{Error analysis for different categories of criteria}
\label{fig:error_analysis}
\end{figure}

We first conducted an error analysis for GPT-4 in generating synthetic data by examining the nature of errors made by the model in its judgments. While a similar analysis can be performed for other models, it is relatively resource-intensive. Thus, we leave such analysis for other models as part of future work.

We manually dived deep into the outputs, particularly focusing on the discrepancies between the model's explanations and the actual classification labels. Our analysis led to the categorization of errors into six distinct types, each highlighting a unique aspect of the model's operational limitations, as shown in Table \ref{tab:error_types_compact}

As shown in Figure \ref{fig:error_analysis}, the majority of mistakes made by the model are around drawing bad judgment when information is not even present in the patient note. For all the inclusion criteria where the annotated output is 'included', the model only makes this type of mistake. For 'no relevant information', however, the majority of mistakes are because the model failed to reason over implicit outputs.

\section{Conclusion and Discussion}
 In this study, we conducted detailed experimental analysis to compare the efficacy of different LLMs, including both open-source and proprietary LLMs, for patient-trial matching. Our observations revealed a significant disparity in the performance of open-source models when juxtaposed with their proprietary counterparts. This divergence in performance is arguably not solely contingent on their pre-training knowledge, but also on their alignment capabilities. As highlighted in \cite{bai2022constitutional}, model alignment plays a pivotal role in achieving superior accuracy across various tasks. In the context of patient matching, which deviates from standard tasks such as question answering or summarization, LLMs struggle to comprehend the task adequately, even when employing one-shot or two-shot methodologies.

\subsection{Can Finetuned Open Source Models Surpass GPT3.5/GPT4?} 

In our comparative analysis, while vanilla open-source LLMs lagged significantly behind GPT-3.5, their fine-tuned counterparts showcased a notable performance leap, not only surpassing GPT-3.5 but also providing strong competition to GPT-4. This enhancement was observed in our study using supervised fine-tuning techniques; however, we posit that incorporating advanced RLHF methods, such as Direct Preference Optimization \cite{rafailov2023direct}, could further elevate the models' capabilities. The improvement was consistent across various evaluation metrics at both the criterion level and the aggregate level. This discovery holds particular relevance for healthcare applications where deploying cost-effective and privacy-conscious AI solutions is crucial. It suggests that fine-tuned versions of open-source models could serve as viable alternatives to GPT-4 for a wide range of use cases. This can help address the challenges associated with higher costs and privacy concerns linked to larger proprietary models.

\subsection{How Much Data Annotation is Needed for Fine-tuning?} 

The process of fine-tuning LLMs presents both computational and methodological challenges, primarily due to the difficulty in providing a dense, multi-token signal that these models require for effective learning. While labeling for classification tasks typically involves single-token signals, enhancing model performance necessitates the provision of multi-token feedback, which is inherently more complex to curate due to its diversity and volume requirements. Despite these challenges, our experiments demonstrate that distillation techniques that have been used to enhance the dialogue capabilities of different models \cite{self, zephyr} can be used for the task of patient matching as well. This method significantly reduces the necessity for manually crafted examples, thereby streamlining the fine-tuning process and making it more affordable.

\subsection{Limitations}
The dataset employed in our experiments, while insightful, does not fully encapsulate the complexity of real-world scenarios, particularly in patient summaries. In practical settings, patient data are often dispersed across various formats, including both structured, coded entries and unstructured notes, sometimes supplemented with imaging data for determining patient eligibility in clinical trials. This multifaceted nature of real-world data, combined with the stringent privacy regulations governing patient information, makes acquiring a truly representative dataset a formidable challenge. As a result, addressing the issues posed by such fragmented data sources is an important area for future work. 

Additionally, it's noteworthy that our annotation process was not manually executed but rather derived from GPT-4 outputs, which could potentially introduce biases or errors affecting model performance. Moreover, we did not follow any sophisticated tools to enforce diversity. To mitigate this, the techniques that we used for curating inclusion and exclusion criteria based on novelty and diversity could be adapted to enhance the diversity of the fine-tuning dataset. Manual curation for a small number of samples could also be used to enhance the performance even further.

While our study did not encompass more recently released models, such as open-source LLMs like Mistral \cite{jiang2023mistral}, or closed-source LLMs like MedPalm 2, which recent research suggests may provide superior performance, we believe this presents another promising direction for future investigation.

\section{Acknowledgments, Funding, Competing Interests}
MN, AB, and HS are employees of Tromics Research.

\bibliographystyle{plain}
\bibliography{sample}
\newpage

\appendix
\section*{Appendix}

\subsection*{Prompts}
We use different prompts for inclusion and exclusion criteria. This was done for two reasons - (a) simplifying the process for the model, and (b) Decreasing the input context length so that models with context length less than 4k can also process the inputs.
\begin{table}[h]
\centering
\begin{tabular}{|p{0.9\linewidth}|} 
\hline
You are an assistant tasked with assessing patient eligibility for clinical trials. Your role involves comparing patient notes with the trial's inclusion criteria, which vary in format and include specifics like age, gender, disease specifics, and medical history. Your responsibilities include:

\begin{enumerate}
    \item Interpreting medical terminology and context in both patient notes and trial criteria.
    \item Explaining the relevance of each inclusion criterion in the patient note, step-by-step.
    \item Annotating relevant sentences from the patient note which are relevant for that criterion or indicating a lack of relevant information.
    \item Labeling each criterion as 'included', 'not included', or 'no relevant information' based on the patient's eligibility for that criterion.
    \item Addressing ambiguities or gaps in patient notes carefully.
    \item Producing a JSON output with the exact format: \texttt{\{"inclusion\_criterion": ["relevance\_explanation", [sentence\_id], "eligibility\_status"]\}} and ensuring its structural accuracy.
\end{enumerate} \\

\hline
\end{tabular}
\caption{Prompt used for generating the Inclusion output}
\label{tab:prompts}
\end{table}

For GPT-4, GPT-3.5 and Base LLAMA models, we provide an example in the following format -

\begin{lstlisting}[language=json]
[
    {
        "role": "system",
        "content" : {prompt}
    },
    {
        "role": "user",
        "content" : "Here is the patient note - {patient_note}. Here is the clinical trial - Title - {title} \n Summary - {summary}\n Target disease - {target_diseases}\n Interventions - {interventions} \n Inclusion Criteria - {inclusion_criteria}
    },
    {
        "role": "assistant",
        "content": "```json```"
    }
]




\end{lstlisting}

We however do not use any example for the fine-tuned models.
\begin{table}[h]
\centering
\begin{tabular}{|p{0.9\linewidth}|}
\hline
You are an assistant tasked with assessing patient eligibility for clinical trials. Your role involves comparing patient notes with the trial's exclusion criteria, which vary in format and include specifics like age, gender, disease specifics, and medical history. Your responsibilities include:

\begin{enumerate}
    \item Interpreting medical terminology and context in both patient notes and trial criteria.
    \item Explaining the relevance of each exclusion criterion in the patient note, step-by-step.
    \item Annotating relevant sentences from the patient note which are relevant for that criterion or indicating a lack of relevant information.
    \item Labeling each criterion as 'excluded', 'not excluded', or 'no relevant information' based on the patient's eligibility.
    \item Addressing ambiguities or gaps in patient notes carefully.
    \item Producing a JSON output with the exact format: \texttt{\{"exclusion\_criterion": ["relevance\_explanation", [sentence\_id], "eligibility\_status"]\}} and ensuring its structural accuracy.
\end{enumerate} \\
\hline
\end{tabular}
\caption{Prompt Used For Generating Exclusion Output}
\label{tab:exclusion_criteria}
\end{table}
\subsection*{Schema}
Ensuring the consistency and parseability of the outputs generated by the models is important for reliable analysis. To achieve this, we implemented a JSON schema against which the output of each model was validated. This approach was particularly important for managing the outputs from LLAMA models, which presented challenges in directly producing JSON-formatted outputs.

The process for handling LLAMA model outputs involved a two-step procedure:
\begin{enumerate}
    \item Extraction of the output text encapsulated between triple backticks (\texttt{\`}) using regular expressions. This step was essential to isolate the relevant output from any extraneous content.
    \item Validation of the extracted output against our predefined JSON schema.
\end{enumerate}

The JSON schema employed for this validation process is as below:
\begin{lstlisting}[language=json]
{
    "$schema": "http://json-schema.org/draft-07/schema#",
    "type": "object",
    "patternProperties": {
        "^.+$": {
            "type": "array",
            "items": [
                {"type": "string"},
                {
                    "type": "array",
                    "items": {"type": "integer"},
                    "minItems": 0,
                    "maxItems": 20
                },
                {"type": "string", "enum": ["included", "not included", "excluded", "not excluded", "no relevant information"]}
            ],
            "minItems": 3,
            "maxItems": 3
        }
    },
    "additionalProperties": false
}
\end{lstlisting}

\subsection*{Examples}

\begin{table}[h!]
\centering
\caption{Example 1 - Trial-LLAMA vs GPT3.5}
\begin{tabular}{ | m{5.5cm} | m{5.5cm} | }
\hline
\multicolumn{2}{|c|}{\parbox{11cm}{\centering 
\begin{enumerate}
    \item A 20-year-old man comes to the emergency due to bleeding after a tooth extraction
    \item The bleeding has persisted for approximately 30 minutes despite constant direct pressure
    \item He is a known case of Hemophilia type A treated with FVIII
    \item Blood pressure is 95/60 mm Hg and pulse is 105/min

    \item  His weight is 70 Kg
    \item Family history is positive for Hemophilia type A in his maternal uncle
    \item  He also has a lipoma on his left arm which he plans to remove surgically
    \item  His FVIII activity is 40%
\end{enumerate}
}
} \\
\hline
\multicolumn{2}{|c|}{FVIII/FIX $<$1\% to 25\%} \\
\hline
\textbf{TrialILLAMA} \faTrophy & \textbf{GPT-3.5} \\
\hline
The patient's FVIII activity is 40\%. & The patient's FVIII activity is 40\%, \textcolor{red}{which falls within the range of $<$1\% to 25\%} \\
\hline
\end{tabular}
\end{table}

\definecolor{darkgreen}{RGB}{0, 100, 0}

\begin{table}[h!]
\centering
\caption{Example 2 - Trial-LLAMA vs GPT3.5}
\begin{tabular}{ | m{5.5cm} | m{5.5cm} | }
\hline
\multicolumn{2}{|c|}{\parbox{11cm}{\centering 
\begin{enumerate}
    \item A 22-year-old Caucasian man came to the Clinic with a history of tremors since a year ago.
    \item The tremor was first in his right hand while holding something.
    \item Later the tremor became continuous and extended to both hands and legs and even at rest.
    \item The Kayser-Fleischer' ring was detected in the ophthalmologic exam.
    \item The physical exam revealed jaundice, hepatosplenomegaly, and hypotonia of the upper limbs.
    \item He had a constant smile on his face, however, he has aggressive behavior according to his parents' explanation.
    \item His laboratory study was significant for a low serum caeruloplasmin (0.
    \item 05 g/l), and a raised 24-hour urine copper excretion (120 $\mu g$/24 h).
    \item Wilson disease was confirmed by high liver copper concentration (305 $\mu g/g$ dry weight of liver).
\end{enumerate}
}
} \\
\hline
\multicolumn{2}{|c|}{\parbox{11cm}{\centering Patients aged 18 to 70 years old requiring a stereotactic robot-guided deep brain stimulation procedure}
}\\
\hline
\textbf{TrialILLAMA} \faTrophy & \textbf{GPT-3.5} \\
\hline
The patient is a 22-year-old man, which falls within the age range of 18 to 70 years old. \textcolor{darkgreen}{However, there is no mention of a requirement for a stereotactic robot-guided deep brain stimulation procedure.} & The patient is 22 years old, within the age range of 18 to 70 years old. \\
\hline
\end{tabular}
\end{table}

\begin{table}[h!]
\centering
\caption{Example 3 - Trial-LLAMA vs GPT4}
\begin{tabular}{ | m{5.5cm} | m{5.5cm} | }
\hline
\multicolumn{2}{|c|}{\parbox{11cm}{\centering 
\begin{enumerate}
    \item A 51-year-old woman is seen in clinic for advice on osteoporosis.
    \item She has a past medical history of significant hypertension and diet-controlled diabetes mellitus.
    \item She currently smokes 1 pack of cigarettes per day.
    \item She was documented by previous LH and FSH levels to be in menopause within the last year.
    \item She is concerned about breaking her hip as she gets older and is seeking advice on osteoporosis prevention.
\end{enumerate}
}
} \\
\hline
\multicolumn{2}{|c|}{\parbox{11cm}{\centering Postmenopausal women or men > 60 years}
}\\
\hline
\textbf{TrialILLAMA} \faTrophy & \textbf{GPT-4} \\
\hline
The patient is a 51-year-old woman \textcolor{darkgreen}{who has been documented to be in menopause within the last year}.& The patient is a 51-year-old woman who is postmenopausal, \textcolor{red}{but she is not older than 60.} \\
\hline
\end{tabular}
\end{table}

\end{document}